# AIwriting: Relations Between Image Generation and Digital Writing

**Scott Rettberg[1], Talan Memmott[2], Jill Walker Rettberg[1], Jason Nelson[1] and Patrick Lichty[2]**

Affiliation (s) [1]University of Bergen, Norway [2]Winona State University

Contact Emails scott.rettberg@uib.no, tmemmott@winona.edu, jill.walker.rettberg@uib.no, jason.nelson@uib.no, patrick.lichty@winona.edu

## Abstract

During 2022, both transformer-based AI text generation systems such as GPT-3 and AI text-to-image generation systems such as DALL•E 2 and Stable Diffusion made exponential leaps forward and are unquestionably altering the fields of digital art and electronic literature. In this panel a group of electronic literature authors and theorists consider new opportunities for human creativity presented by these systems and present new works have produced during the past year that specifically address these systems as environments for literary expressions that are translated through iterative interlocutive processes into visual representations. The premise that binds these presentations is that these systems and the works generated must be considered from a literary perspective, as they originate in human writing. In works ranging from a visual memoir of the personal experience of a health crisis, to interactive web comics, to architectures based on abstract poetic language, to political satire, four artists explore the capabilities of these writing environments for new genres of literary artist practice, while a digital culture theorist considers the origins and effects of the particular training datasets of human language and images on which these new hybrid forms are based.

## Keywords

AI, GPT-3, DALL•E 2, Stable Diffusion, electronic literature, image generation, digital narrative, language models

## Patrick Lichty: Latent Space

Aesthetics of the Latent Space: Taxonomies and Architectures of the Latent Space will present a durational exploration of the poetics of Machine Learning-based image generation. Text-based machine learning has swept the digital art world from generative PFPs (Bored Ape Yacht Club) to poetic AI-based art exhibitions (The Grand Exhibition of Prompts). For the past four years, the author has been exploring the aesthetic of the latent space in AI-based imaging systems. Examining these image spaces centers on the methodological exploration of GANs and CLIP-based prompt-based art. As opposed to the representational figurative and landscape work commonly seen, the two works discussed explore analytical spaces designed to explore the poetics of Machine Learning. The first, "Personal Taxonomies," stems from a GAN-based analysis of over one thousand daily abstract calligraphies compared to finding Chomsky-Esque "deep structures" of commonality between the glyphs. Conversely, "Architectures" seeks to create ambiguous spaces subtly adjusted when representation becomes too explicit about exploring the visual locus of the latent space.

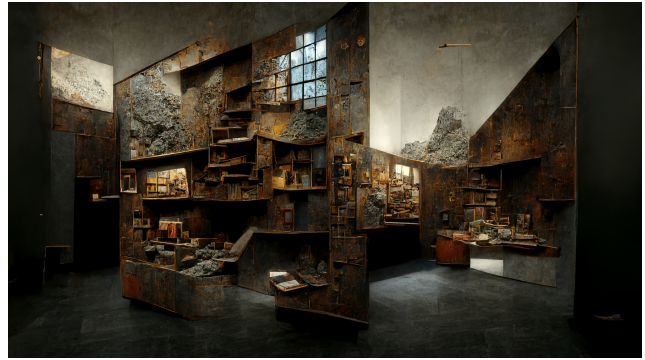

Figure 1. Image from "Architectures of the Latent Space." CC-BY 2022 Patrick Lichty

As the author wrote in the latest volume of Hegeland's "The Future of Text" anthology [1], the creation of prompt-based Machine learning-generated images is not about art but the intersection of code and poetics and visual concretism. In the case of Personal Taxonomies, this arises from a latent space of images based on Foucault's notion of the Calligram or the liminally legible glyph. AI/GANs are used to differentiate/decode these images into forms of personal Rorschach. Conversely, Architectures focuses on the prompt as a site of abstract concretized prose, with the writer struggling against representation. This presentation compares/contrasts these projects and examines the formal/aesthetic relations between AI models and their relation to the poetics of machine learning-based imaging.

## Jason Nelson: Prompt and Rethink

"The Awkward Handshake," "The many occasions of moving," and "robot birds are not" are all works/collections of digital writing that engage with our utilize AI tools/engines/code/systems in some manner. Albeit most iterations of digital art engage with machine intelligence in some form, with intelligence being the ideological hinge. What seems to separate AI systems from those with more iconic/symbolic interfaces (the arrow selects, the glass magnifies) is the conversation between the human creator and our code-made collaborators. These three works engage with this conversation, this prompt and change and rethink process, in variant ways, divergent for the creator and what and how the user/player/reader experiences.

"The Awkward Handshake" explores how to create the programmable skeleton of digital writing with AI tools. With coaxing and persistence, the exchange between GTP3 and my hands cobbles together a series of interfaces for digital writing. The conversation to create these literary codesets is truncated and clumsy, as the AI cannot (or will not) experience the interaction, cannot move the words and click through the poetic amalgamation of image and word. The result is a series of guesses by both actors, with the digital poet as director and arbitrator and eventual claimant of authorship.

What's spoken between the human and the AI in "The many occasions of moving" is less about the artwork and more about inspiration for the creation. As a digital art-game, TMOM creates a series of poetic worlds the reader/player can inhabit and explore. The visuals are derived from text-to-image AI chatting and cutting and reforming. The detailed imagery living in the work was only possible through this conversation. The detail of shading and style, the accurate imperfections of line and object, are the product of the engine scrapping the creativity of giant hoards of humans. That's very helpful for a digital poet with poor drawing skills.

Whereas the first two works are built from mining the expertise inherently stacked in the AI corpus, "robot birds are not" switches the human writer role to one of translation. After asking the programmed brain to make a series of surreal, broken, but recognisably comic strip, images, I attempt to understand the narrative created in the visuals. I write the bubbling speech, create the narrative connections between frames, give the comic strip images lives and relational fun-times. This process continues, two electric brains, bouncing ideas. The result is a graphic novel, built in the back and forth and imagined during hundreds of loading bars.

Nevertheless I find it critical to understand that while I imagine having a conversation with these AI tools, and I imagine we are collaborating to create digital writing, interactive and visual narratives, the AI does not know or care. I am hoping to birth new digital creatures, whereas my programmed other has no interest either way. I can type or close, listen or leave, ask or mumble, but the maybe smart computer only borrows from what the world has made, passes it on to me in variations, uncaring and yet, at times, disarmingly beautiful.

**Talan Memmott: Reinventing the Self**

"Introducing Lary" is an ongoing AI project that explores the re-invention of the self following life-changing cancer surgery and treatment. Based on the artist's own experience with a laryngeal cancer diagnosis and the medical interventions that follow, the project uses a variety of AI image generating platforms along with journal entries and medical reports as text prompts to produce images that range from the mythic to the horrific; the historical to speculative.Though at times the resulting images may emphasize the emergent body horror of cancer surgeries and treatments, they serve as a form of therapeutic aesthetics for the artist as patient.

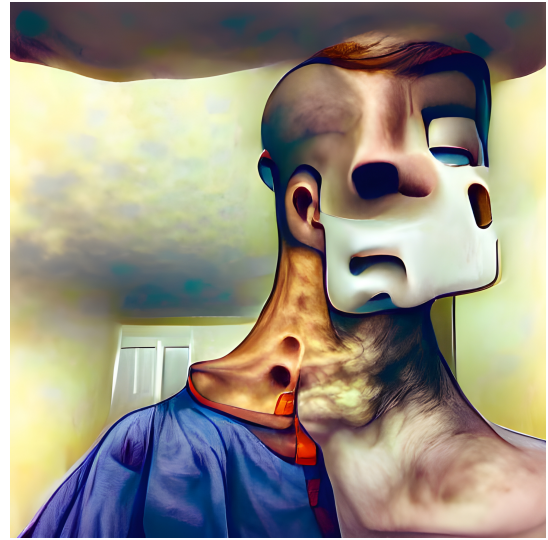

Figure 3. "Stoma" from "Introducing Larry". CC-BY 2022 Talan Memmott

Rather than allowing the images to remain purely digital artifacts, the work extends its output in physical forms such as large scale canvas prints that include interventions such as tracheotomy tubes and sutures, and photo books that include poetic and narrative texts. "Alocutive Interpolation," a 1.5 by 1.5 meter canvas print, the first physical manifestation of the "Introducing Lary " series, was exhibited at the International Digital Media Arts Association Weird Media Exhibition in June of 2022.

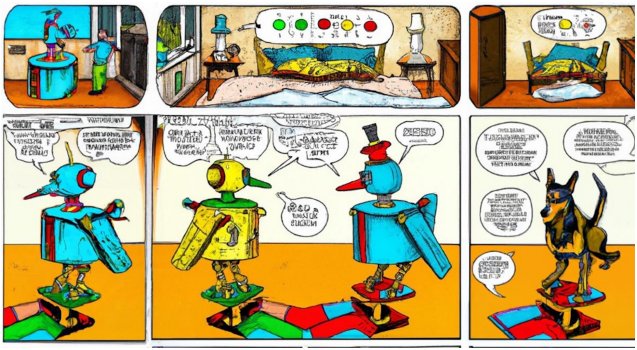

Figure 2. Image from "The Awkward Handshake". CC-BY 2022 Jason Nelson

## Scott Rettberg: Text-to-Image Political Parody

"Republicans in Love" is an AI text-to-image project produced during the month following the November 2022 United States Congressional election that explores the extent to which platforms such as DALL•E 2 can be used for satirical literary purposes. The project brings together elements of art history, politics, and social media discourse while also serving as a case study in the capabilities and limitations of the platform. "Republicans in Love" is a series of images based on one-line prompts that revisit historical incidents, ironies, and dangers of contemporary Trumpian populism. At the same time, the project traverses the history of European and American visual art through the manifestation of the styles of artists specified in the prompts. The project has resulted in an artist's book and a set of playing cards and will be featured in an exhibition of prints in Bergen in 2023.

How should we to understand the function and ontological status of text-to-image generation systems such as DALL•E 2, Stable Diffusion, Midjourney AI and so on? While some fear that these systems are either harbingers of doom for human artists and designers or techno-utopian manifestations of the achievement of a foundational model of generalized AI, it is the author's premise that from the user's perspective, AI-based text-to-image generation systems are best understood as writing environments. The human writer engages with the nonconscious cognitive system of the AI, which accesses immense datasets of human language and existing imagery. The system attempts to draw correlations between the language provided the user, approximations of where those language elements might meet in the latent space of its language dataset, and corresponding image elements that can be understood as conceptual approximations of the language provided. For example, a prompt such as "Republicans in love, angry about the news, eating greasy cheeseburgers at the President's desk in the Oval Office, in the style of Caravaggio" will yield images that to varying degrees of success draw together many of the elements provided.

The resulting image clearly incorporate some elements of the Renaissance painter's style, represents an idea of what a "Republican" might look like, integrates the burger, interior design similar to that of the oval office, and deals fairly well with the seemingly contradictory emotions of love and anger. There are many different ways of interpreting what this AI system is ontologically. Characterizing the interaction between human cognition, writing, machine cognition, and image production to the status of an artist's "tool" would be reductive. Rather than conceptualizing this process as akin to that of producing an image with Photoshop, we should understand it as an environment for the literary production of visual narrative. The results of Latent Diffusion Models, or Stable Diffusion systems in this regard bear a strong resemblance to practices in postmodern literature and visual art, as they function as pastiche machines, blenders of immense proportions that access to massive datasets of human expression, bringing together semiotic vertices that the systems can calculate and sketch but not comprehend. The collection of approximately 100 text and image pairs of "Republicans in Love" serves an experiment in using this environment of human and machine cognition to produce a sustained and recognizably literary work.

Figure 4. "Republicans in love, angry about the news, eating greasy

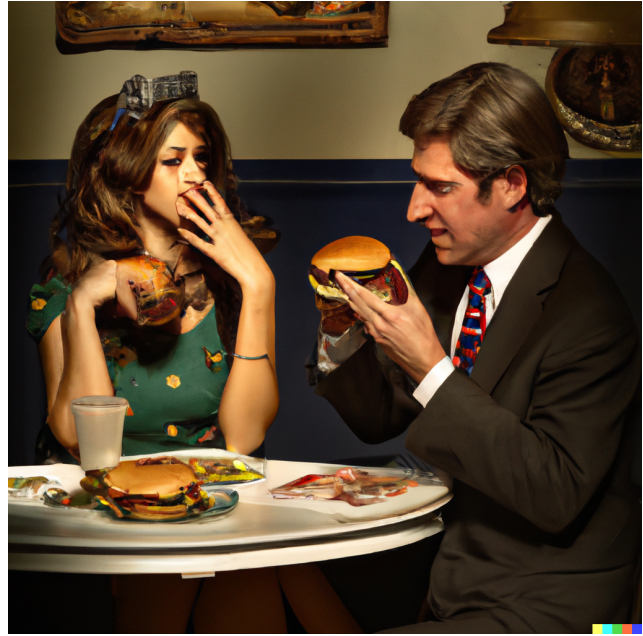

cheeseburgers at the President's desk in the Oval Office, in the style of Caravaggio" from "Republicans in Love". CC-BY 2022 Scott Rettberg

## Jill Walker Rettberg: Dataset Culture

How does the data that AI models are trained on affect the art and literature they able to generate? This paper uses a critical dataset studies approach to identify the data current AI models are built upon. It then analyses selected output to demonstrate how certain genres and styles of art are emphasised above others.

Generative AI models like DALL-E and the GPT series are trained on massive datasets of images and texts. They are trained on our shared cultural heritage - or at least on a small part of it. In fact, the GPT models are mostly trained on websites that have been upvoted on Reddit, self-published fiction and the English-language Wikipedia (Brown et.al. 2020). Audits of the datasets have found that more than half the websites are from US domains, and that the countries with the 2nd, 3rd and 4th largest English speaking populations (India, Pakistan, Nigeria and the Philippines) are represented in less than 4% of the corpus. Any text with words on a blocklist meant to avoid porn, swearwords and slurs is filtered out, but this also disproportionately removes writing in minority versions of English (e.g. African American or Hispanic English) and queer content (Dodge et.al. 2021). Image generation models are trained not only on images, but also using semantic structures like ImageNet, which Trevor Paglen and Kate Crawford's art project *ImageNet Roulette* (2019) and their accompanying paper clearly demonstrated has deep structural biases. AI bias has received a lot of well-earned criticism. But when we use AI to create fiction, art and literature, we may not want to filter out all the flirtation, insults, threats and sexual references that are defined as "toxic".

By processing billions of words and images, AI models build models of human culture. What is literature for an AI? What is art? What do billions of words and images tell us about literature? To begin to find out, I have used ChatGPT to generated plot summaries for different literary genres using prompts like "Write a plot summary for a science fiction novel", "Write code to generate an interactive fiction", or "Write a sonnet about winter". Other variants add themes and topics. By comparing the output, I will identify genre characteristics that can be compared to those of human-authored writing. A final discussion will examine how the training data can be said to relate to the literary output of the AI model.

## Authors Biographies

**Patrick Lichty** is a media "reality" artist, curator, and theorist who explores how media and mediation affect our perception of our environment. He is best known for his work as a principal of the virtual reality performance art group Second Front, and the animator of the activist group, The Yes Men. He is a CalArts/Herb Alpert Fellow and Whitney Biennial exhibitor as part of the collective RTMark. His recent book, Variant Analyses: Interrogations of New Media Culture was released by the Institute for Networked Culture, and is included in the Oxford Handbook of Virtuality.

**Talan Memmott** is a digital writer/artist/theorist. Memmott has taught and been a researcher in digital culture and media practices at University of California Santa Cruz; University of Bergen, Norway; Blekinge Institute of Technology in Karlskrona, Sweden; California State University Monterey Bay; the Georgia Institute of Technology; and, the University of Colorado Boulder. He is Associate Professor of Creative Digital Media at Winona State University. Memmott holds an MFA in Literary Arts/Electronic Writing from Brown University and a PhD in Interaction Design/Digital Rhetoric and Poetics from Malmö University.His digital art and electronic literature work has been exhibited, presented, and published internationally. He was a co-editor for the Electronic Literature Collection, Volume 2 (ELO), the ELMCIP Anthology of European Electronic Literature, and was the recipient of the 2021 Electronic Literature Organization Maverick Award.

**Jason Nelson** is a creator of curious and wondrous interactive poems and fictions of odd lives, builder of confounding art games and all manner of curious digital creatures. He professes Digital Narrative at the University of Bergen in Norway. Aside from coaxing his students into breaking, playing and morphing their creativity with all manner of technologies, he exhibits/publishes widely in galleries and journals, with work featured at ARS, FILE, ACM, LEA, ISEA, SIGGRAPH, ELO and other acronyms. There are awards to list (Paris Biennale Media Poetry Prize, Digital Writing Award, New Media Writing Prize), and Fellowships he's adventured into (Fulbright, Moore), but it's the Webby award that makes him smile. play more at:  dpoetry.com

**Scott Rettberg** is Director of the Center for Digital Narrative and Professor of digital culture in the Department of linguistic, literary, and aesthetic studies at the University of Bergen, Norway and is Director of the Center for Digital Narrative. He is the author or coauthor of novel-length works of electronic literature, combinatory poetry, and films including *The Unknown, Kind of Blue, Implementation, Frequency, The Catastrophe Trilogy, Three Rails Live, Toxi\*City, Hearts and Minds: The Interrogations Project* and others that have been exhibited widely in digital art venues around the world. He is the cofounder of the Electronic Literature Organization. Scott Rettberg's book *Electronic Literature* (Polity, 2018) was the winner of the 2019 N. Katherine Hayles Award for Criticism of Electronic Literature.

**Jill Walker Rettberg** is Professor of digital culture at the University of Bergen in Norway and Principal Investigator of the ERC project "MACHINE VISION: Machine Vision in Everyday Life: Playful Interactions with Visual Technologies in Digital Art, Games, Narratives and Social Media." Rettberg's book *Machine Vision: How Algorithms are Changing the Way We See the World* is forthcoming on Polity Press in 2022. Her previous book, *Seeing Ourselves Through Technology: How We Use Selfies, Blogs and Wearable Devices to See and Shape Ourselves*, was published as an open access monograph by Palgrave in October 2014, and can be freely downloaded. Her book *Blogging* was published in a 2nd edition by Polity Press in 2014, and she has also co-edited an anthology of critical writing on World of Warcraft (MIT Press 2008).